\documentclass{article}

\usepackage{microtype}
\usepackage{graphicx}
\usepackage{subfigure}
\usepackage{booktabs} 
\usepackage{url}  

\usepackage{bm}
\usepackage{enumitem}
\usepackage{pifont}
\usepackage{amssymb}
\usepackage{amsmath}
\usepackage{multirow}
\usepackage{hyperref}

\usepackage[accepted]{icml2019}

\icmltitlerunning{LatentGNN: Learning Efficient Non-local Relations for Visual Recognition}

\begin{document}

\twocolumn[
\icmltitle{LatentGNN: Learning Efficient Non-local Relations for Visual Recognition}





\begin{icmlauthorlist}
\icmlauthor{Songyang Zhang}{to}
\icmlauthor{Shipeng Yan}{to}
\icmlauthor{Xuming He}{to}
\end{icmlauthorlist}

\icmlaffiliation{to}{School of Information Science and Technology, ShanghaiTech University, Shanghai, China}

\icmlcorrespondingauthor{Xuming He}{xmhe@shanghaitech.edu.cn}

\icmlkeywords{Machine Learning, ICML}

\vskip 0.3in
]



\printAffiliationsAndNotice{}  

\begin{abstract}
Capturing long-range dependencies in feature representations is crucial for many visual recognition tasks. Despite recent successes of deep convolutional networks, it remains challenging to model non-local context relations between visual features. A promising strategy is to model the feature context by a fully-connected graph neural network (GNN), which augments traditional convolutional features with an estimated non-local context representation. However, most GNN-based approaches require computing a dense graph affinity matrix and hence have difficulty in scaling up to tackle complex real-world visual problems. In this work, we propose an efficient and yet flexible non-local relation representation based on a novel class of graph neural networks. Our key idea is to introduce a latent space to reduce the complexity of graph, which allows us to use a low-rank representation for the graph affinity matrix and to achieve a linear complexity in computation. Extensive experimental evaluations on three major visual recognition tasks show that our method outperforms the prior works with a large margin while maintaining a low computation cost. For facilitating the future research, code is available: \url{https://github.com/latentgnn/LatentGNN-V1-PyTorch}

\end{abstract}

\section{Introduction}

Modeling non-local relations in feature representations is a fundamental problem in visual recognition, which enables us to capture long-range dependencies between scene entities and handle feature ambiguity for high-level semantic understanding tasks~\cite{strat1991context}. In particular, tremendous progress has been made recently in computer vision community by learning hierarchical feature representations of images with deep convolutional networks. Nevertheless, the dominating feature learning strategy in deep Convnets is through stacking many local operations of convolution and pooling, which are effectively limited to local receptive fields and short-range context~\cite{luo2016understanding,chen2017rethinking}. 


There have been several attempts in the literature to tackle the problem of incorporating non-local context into representations of existing deep models. Early works tend to incorporate recurrent neural network (RNN) module to capture the long-range dependencies by constructing a directed acyclic graph model in convolutional feature space~\cite{byeon2015scene,shuai2018scene} or semantic label space~\cite{zheng2015conditional}. These RNN-based models are typically defined on graphs with local neighborhood or specific edge functions, which are limited in modeling non-local relations and computationally expensive. An alternative strategy of context modeling is to exploit image structure and integrate a region-based representation with deep convolutional features~\cite{li2018beyond}. Regions and their relations (e.g., similarity) are typically modeled as an undirected graph, which can be learned jointly with convolutional features to encode long-range dependencies. The region generation~\cite{gould2009region}, however, relies on bottom-up grouping that is sensitive to low-level image cues and may yield erroneous representations.     


More recent efforts have been focused on introducing non-local operations into the deep Convnets~\cite{wang2018non,yue2018compact}. In particular, \citet{wang2018non} propose to augment the convolutional features by a context representation computed from the entire feature map at certain convolution layers. Such non-local operations can be viewed as a densely-connected graph neural network (GNN) module with attention mechanism~\cite{velickovic2017graph}, which estimates the feature context by a single iteration of message passing on the underlying graph.  
This feature augmentation method allows a flexible way to represent non-local relations between features and has led to significant improvements in several vision recognition tasks~\cite{wang2018non}. However, despite its successes, the GNN-based non-local neural network suffers from two major limitations, especially in the context of visual recognition. First, the graph message passing requires computing a dense affinity matrix of the graph and its product with the node messages, which are computationally prohibitive for large-sized feature maps. In addition, the graph edge affinities, which capture relations between two node features, are typically defined by a simple pairwise function (with possible normalization) and hence has a limited capacity in modeling complex relationships.       



In this work, we propose an efficient and yet flexible non-local relation representation of convolutional features to address the aforementioned limitations. Our goal is to achieve scalable, fast message passing and low-cost context-aware feature augmentation for visual recognition tasks. 
To this end, we develop a novel graph neural network module, inspired by the latent variable graphical models~\cite{koller2009probabilistic}, for encoding long-range dependency between features. Specifically, instead of propagating context information on a fully-connected graph, we introduce an augmented graph with a set of latent nodes that are connected to convolutional features and themselves. Typically the dimension of the latent space is much lower than the convolutional features, which enable us to design an efficient message passing procedure to compute non-local context information for feature nodes through the `global' latent space. Figure.~\ref{fig:message_passing} shows an overview of our model structure and inference process. 

Our message passing algorithm consists of three steps in a latent graph network: each latent node first collect messages from convolutional feature nodes, which is followed by message propagation among latent nodes, and finally the messages from latent nodes are propagated back to the feature nodes. Mathematically, our algorithm is equivalent to representing the graph affinity matrix of the fully-connected feature graph as a mixture of low-rank kernel matrices defined on convolutional features. Such equivalence allows us to introduce a parametrized mixture of low-rank matrices to encode a rich set of non-local relations and an end-to-end task-driven training strategy to learn the relations and feature representation jointly. Moreover, we can easily stack multiple layers of the basic latent graph network modules in any deep ConvNet architecture.      



We validate our model on two typical classes of feature graphs in vision tasks: the first is grid-like convolutional feature graphs that are widely adopted in object detection and instance segmentation; and the second is irregular-shaped feature graphs for point cloud semantic segmentation. Extensive experiment evaluations show that our model outperforms the baselines with a large margin while maintaining a low computation cost, indicating its broad applicability in a variety of vision tasks.
 
The main contributions of this work are summarized as the following:
\begin{itemize}
 \setlength{\itemsep}{0pt}
 \setlength{\parskip}{0pt}
 \setlength{\parsep}{0pt}
	\item We propose an efficient feature augmentation framework to incorporate global context and long range dependency through a novel graph neural network for visual recognition tasks.
	\item We develop a flexible representation for non-local relations based on mixture of low-rank kernel matrices for fast message passing, which can be learned jointly with convolutional features with task-aware losses.
	\item Our model has a modularized design, which can be easily incorporated into any layer in deep ConvNets, and stacked into multiple blocks with different computation cost constraints.
\end{itemize}

\section{Related Work}\label{sec:related}

\paragraph{Context-aware Representation Learning}
While deep convolution networks are capable of extracting a hierarchical feature representation from an image, the effective receptive fields of their neurons are typically limited to local regions and unable to capture long-range dependencies~\cite{luo2016understanding,chen2017rethinking}.



There have been a large body of works aiming to incorporate context information into existing deep ConvNet models. One type of strategies tends to aggregate context information through the image pyramid techniques. \citet{he2014spatial} propose to use 'spatial pyramid pooling' to aggregates local features for object detection. \citet{chen2018deeplab} introduce atrous spatial pyramid pooling with multi-scale dilation convolution for richer contextual information. \citet{zhao2017pyramid} introduce pyramid pooling to capture feature context. However, these pyramid pooling-based methods typically reply on homogeneous contextual information and lack the capacity to capture pixel-wise relationship. Enlarging the receptive field explicitly is also a common strategy. \citet{yu2015multi} explore dilated convolution for large receptive filed, and \citet{dai2017deformable} improve the standard convolution operator to handle the deformation and varying object scales. However, these methods are still unable to capture non-local relations as the receptive fields mostly cover local neighborhood and short-range contextual information. 

Region-baed methods exploit image regions for context reasoning and visual recognition. \citet{gould2009region} propose a hierarchical region-based approach to reason about pixels, regions and objects in a coherent probabilistic model. \citet{russell2009associative,li2018beyond} introduce flat graph structures--with regions as vertices and the similarity between regions as edges for recognition and context modeling. However, region generation is challenging but essential in this paradigm, which requires careful tuning and extra efforts to design.



\paragraph{Graph Neural Networks}
Our work is related to learning deep networks on graph-structured data. The GNN was first proposed in \cite{gori2005new,scarselli2009graph}, which is a trainable recurrent message passing process. Before the GNN, graphical models can also model the long-range dependencies such as the conditional random field~\cite{he2004multiscale}, and it are exploited as the post-processing in the context of deep Convnets~\cite{chen2018deeplab,zheng2015conditional}. Graph neural network are natural generalizations of convolutional neural networks to non-Euclidean graphs. \citet{bruna2013spectral} and \citet{henaff2015deep} propose to learn smooth spectral multipliers of the graph Laplacian. \citet{kipf2017semi} introduce learning polynomials of the graph laplacian instead of compute eigenvectors to alleviate the computational bottleneck. 
\citet{kearnes2016molecular,gilmer2017neural} generalize the GNN and also learn edge features from the current node hidden representations.



\paragraph{Latent Variable Models and Kernel Learning}

The idea of using latent variables to capture long-range dependency has been widely explored in the literature of probabilistic graphical model~\cite{koller2009probabilistic}. Our graph structure resembles the Restricted Boltzmann Machines with lateral connections~\cite{osindero2008modeling}. However, we mainly use the latent graph for feature computation instead of defining a distribution over random variables.

Our work is also inspired from low-rank approximation in kernel methods~\cite{scholkopf2002learning}. One limitation of kernel methods is their high computation cost, which is at least quadratic in the number of training examples, duet to the kernel matrix calculation. To avoid computing kernel matrix, one common approach is to based on approximate kernel learning, which generates a vector representation of data that approximates  kernel similarity between any two data points. The most well known approaches in this category are \textit{random Fourier features} \cite{rahimi2008random,rahimi2009weighted} and \textit{the Nystr\"om method} \cite{williams2001using,drineas2005nystrom}. Rondom Fourier features is an approach to scaling up kernel methods for shift-invariant kernels. Nystr\"om method that approximate the kernel matrix by a low rank matrix.

\paragraph{Visual Recognition Tasks}
Visual recognition community has witnessed tremendous development with deep convolutional neural network. Object detection, instance segmentation, point cloud semantic segmentation and many other challenging visual recognition tasks have been actively studied in plenty of recent works. 

For image object detection, a common branch is R-CNN \cite{girshick2014rich} based methods, which is detecting objects based on region proposals. Fast R-CNN \cite{girshick2015fast} was proposed to reduce the computation cost of R-CNN by extracting the whole image feature onceRoI Align layer \cite{he2017mask} is designed to address the coarse spatial quantization for improving RoI pooling further. Faster R-CNN \cite{ren2015faster} introduce a region proposal network(RPN) that share the same backbone with the detection to build the first unified end-to-end efficient framework for object detection.

Deep learning with 3D data is a growing filed of research recently. There a series of works propose several effective architectures to process point cloud directly. PointNet \cite{qi2017pointnet} is a pioneering effort to handler the unordered point data, which applies deep learning to point clouds by point-wise encoding and aggregation through global max pooling. To capture more local geometric details, PointNet++ \cite{qi2017pointnet++} introduce a hierarchical neural network to learn multi scale feature descriptor for the point cloud. PointCNN \cite{li2018pointcnn} propose $\mathcal{X}$-conv layer instead of MLP layer to exploit canonical ordering of points. We conduct experiments both on grid-like data irregular-shaped data to validate  our method is generic and effective. Superpoint Graphs \cite{landrieu2018large}(SPG) first partitions point cloud into superpoints and encode every super point with PointNet. The semantic labels of superpoints are predicted from the PointNet embedding of current superpoint and spatially neighboring superpoints. 

\section{Our Approach}\label{sec:model}

Our goal is to learn a context-aware feature representation for a visual input (e.g., image or point cloud) that captures long-range dependencies between convolutional features. To this end, we adopt a graph neural network (GNN) framework to model the non-local relations of conv-features, and develop an efficient message passing algorithm to estimate non-local context representation of each conv-feature. Unlike previous non-local relation networks~\cite{wang2018non,yue2018compact}, our strategy proposes a novel graph structure for fast message updates and a flexible graph affinity representation for learning the non-local relations in a task-driven manner.

Below we start with a brief introduction to the GNN framework~\cite{wang2018non,velickovic2017graph} for estimating non-local context information and motivate our network design in Sec.~\ref{subsec:overview}. We then present our network structure and message passing algorithm in Sec.~\ref{subsec:latent}, followed by a low-rank kernel learning formulation in Sec.~\ref{subsec:kernel}. Finally, we describe feature augmentation strategies that combine our GNN module with deep ConvNets for visual recognition tasks in Sec.~\ref{subsec:incorporate}. 

\begin{figure*}[!th]
	\centering
	\includegraphics[width=1.0\linewidth]{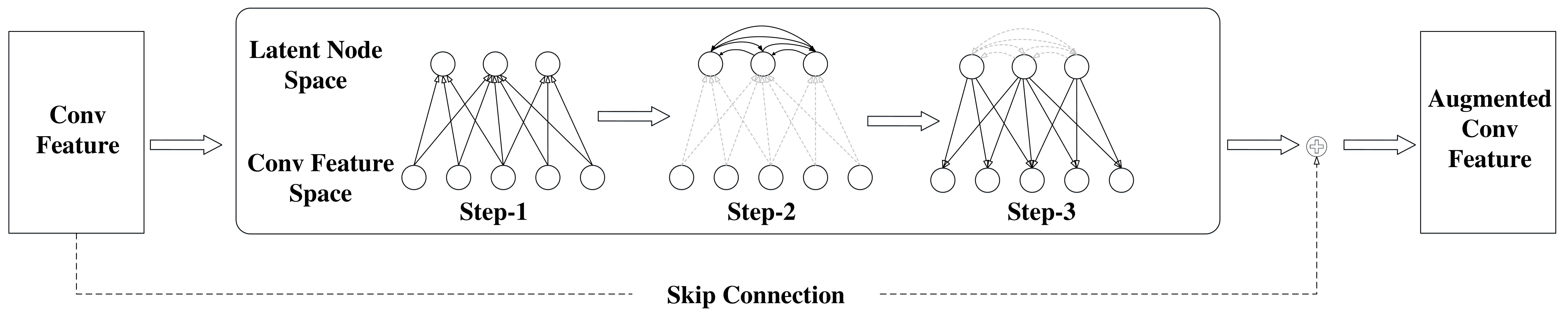}
	\caption{Illustration of the latent graph network. Step-1 shows  the message passing from convolutional feature space the latent node space, and message passing between pairwise nodes in latent node space is step 2. Step-3 indicates propagated message back to the convolutional feature nodes. }
	\label{fig:message_passing}
\end{figure*}

\subsection{Non-local Graph Neural Networks}\label{subsec:overview}


We adopt the formulation in~\cite{wang2018non} for the non-local context feature estimation and first introduce the fully-connected graph neural network for this task. Formally, let $\mathbf{X} = [\mathbf{x}_1,\cdots, \mathbf{x}_N]^\intercal$ be a set of convolutional features, where $\mathbf{x}_i\in\mathbb{R}^c$ is a $c$-channel feature vector and $i$ indexes the spatial location of the feature vector. Taking image features as an example, we have a conv-feature map defined on a 2D grid with the size of $c\times h\times w$, where $h$, $w$ and $c$ are the height, width and number of channels, respectively. Here $N=w\cdot h$ and for $\mathbf{x}_i$, $i$ indexes the spatial location on the feature map. 

We then build a fully-connected graph $\mathcal{G}=(\mathcal{V},\mathcal{E})$ with $N$ nodes $v_i\in \mathcal{V}$,  and edges $(v_i, v_j) \in \mathcal{E}, \forall i<j$, to represent the non-local relations between features. 
The graph neural network assigns $\mathbf{x}_i$ as the input to the node $v_i$ and computes the feature representation of each node through an iterative message passing process. Each iteration updates the node features as follows, 
\begin{align}
	\mathbf{\tilde{x}}_i &= h \left(\frac{1}{Z_i(\mathbf{X})} \sum_{j=1}^N g(\mathbf{x}_i,\mathbf{x}_j)\mathbf{W}^{\intercal}\mathbf{x}_j\right)\label{eq:graph_aug}
\end{align}
where $\mathbf{\tilde{x}}_i$ represents the updated feature representation at node $i$, $h$ is an element-wise activation function (e.g., ReLU). $g$ is a kernel function encoding relations between two feature vectors $\mathbf{x}_i$ and $\mathbf{x}_j$, and $Z_i(\mathbf{X})$ is a normalization factor for node $i$. $\mathbf{W}\in \mathbb{R}^{c\times c}$ is the weight matrix defining a linear mapping to encode the message from node $i$. 
Note that we can write the updating equation Eq~\eqref{eq:graph_aug} in a matrix form:
\begin{align}
\mathbf{\tilde{X}} = h\left(\mathbf{A}(\mathbf{X})\mathbf{X}\mathbf{W}\right)\label{eq:matrix_mp}
\end{align}  
where $\mathbf{A}(\mathbf{X})\in \mathbb{R}^{N\times N}$ denotes the affinity matrix of the graph in which $\mathbf{A}_{i,j}=\frac{1}{Z_i(\mathbf{X})}g(\mathbf{x}_i,\mathbf{x}_j)$. Moreover, we can easily extend this updating procedure to multiple iterations by unrolling the message passing into a multi-layer network~\cite{gilmer2017neural}. We will focus the single iteration setting in the remaining of this section for notation clarity.

As the updated $\mathbf{\tilde{x}}_i$ integrates context information from the entire feature set $\mathbf{X}$ through message passing, we can use it as an estimate of the non-local context for feature $\mathbf{x}_i$. For visual recognition, the common strategy for defining the graph affinity matrix is based on similarity of neighboring node features (denoted as $\mathbf{A}_{\text{sim}}$) or the random walk graph Laplacian of node features (denoted as $\mathbf{A}_{\text{lap}}$). Specifically, for $\mathbf{A}_{\text{sim}}$, we define $g(\mathbf{x}_i,\mathbf{x}_j)=\mathbf{x}_i^\intercal\mathbf{x}_j$ and $Z_i(\mathbf{X})=1$, while for $\mathbf{A}_{\text{lap}}$, we have the same $g$ function but set $Z_i(\mathbf{X})=\sum_j g(\mathbf{x}_i,\mathbf{x}_j)$. These two graph affinity matrices can be written compactly as follows,
\begin{align}
\mathbf{A}_{\text{sim}}&=\mathbf{M}=\mathbf{X}\mathbf{X}^\intercal \\
\mathbf{A}_{\text{lap}}&=\mathbf{D}^{-1}\mathbf{M}
\end{align}
where $\mathbf{M}\in \mathbb{R}^{N\times N}$, and $\mathbf{D}$ is the diagonal degree matrix in which $\mathbf{D}_{ii}=\sum_j\mathbf{M}_{ij}$.

While the non-local graph neural networks are capable of effectively capturing long-range dependency in conv-features, they suffer from two inherent limitations in the context of visual recognition. First, the message passing in Eqn~\eqref{eq:matrix_mp} involves computing the matrix product $\mathbf{A}(\mathbf{X})\mathbf{X}$, which has a quadratic complexity in terms of size of the graph ($\mathcal{O}(N^2\cdot c)$). Hence this 
inference process is computationally prohibitive for large-sized feature graphs, which restricts its usage in real-world vision tasks. Moreover, the graph edge affinities, which capture relations between two node features, are typically defined by a simple pairwise function (with possible normalization) and hence has a limited capacity in modeling complex relationships. To address those issues, we propose a new graph neural network model in the following subsections.

\subsection{Latent GNN for Non-local Feature Context}\label{subsec:latent}

Instead of relying on fully-connected graph defined on input features, we now introduce a new graph structure to efficiently encode the long-range dependencies. To achieve this, we first augment the original feature nodes with a set of additional latent nodes and then connect those nodes in a structural manner. Our design is inspired by the undirected graph models with latent representations, such as the Restricted Boltzmann Machine (RBMs)~\cite{nair2010rectified}. The newly-designed neural network is capable of encoding non-local relations through a shared latent space and admits efficient message passing due to its sparse connections. 

Specifically, we introduce a set of latent features $\mathbf{Z} = [\mathbf{z}_1,\cdots, \mathbf{z}_d]^\intercal$ where $\mathbf{z}_i\in\mathbb{R}^c$ is a $c$-channel feature vector and the number of latent features $d\ll N$. We then build an augmented graph $\mathcal{G}_L=(\mathcal{V}_L,\mathcal{E}_L)$ with $N+d$ nodes, $\mathcal{V}_L= \mathcal{V}\cup\mathcal{V}_h$. Here $\mathcal{V}$ corresponds to the input conv-features and $\mathcal{V}_h$ is associated with the latent features. We also define the graph connectivity by two subsets of graph edges, $\mathcal{E}_L=\mathcal{E}_v\cup\mathcal{E}_h$, where $\mathcal{E}_v$ denote the connections between $\mathcal{V}$ and $\mathcal{V}_h$ while $\mathcal{E}_h$ are the graph edges within $\mathcal{V}_h$.     

Our latent graph neural network takes the conv-features $\mathbf{X}$ as inputs to the nodes in $\mathcal{V}$ and computes context-aware representations of each node through an iterative message passing with a specific scheduling scheme. Concretely, each iteration consists of three steps for updating the node features (see Figure.\ref{fig:message_passing} for the illustration):

Step 1. \textit{Visible-to-latent propagation.} We update the latent features by propagating messages from the conv-feature nodes to the latent nodes:
\begin{align}
\mathbf{z}_k = \sum_{j=1}^{N}\psi(\mathbf{x}_j,\theta_k)\mathbf{W}^\intercal\mathbf{x}_j, \quad 1\leq k \leq d,\label{eq:up}
\end{align} 
where $\psi(\mathbf{x}_j,\theta_k)$ encodes the affinity between the feature node $\mathbf{x}_j$ and the latent node $\mathbf{z}_k$, and $\theta_k$ is the parameter of the affinity function.  

Step 2. \textit{Latent-to-latent propagation.} We refine the latent features by propagating messages in the fully-connected latent subgraph $(\mathcal{V}_h,\mathcal{E}_h)$:
\begin{align}
\mathbf{\tilde{z}}_k = \sum_{j=1}^{d}f(\phi_k,\phi_j,\mathbf{X})\mathbf{z}_j,\quad 1\leq k \leq d,\label{eq:hid}
\end{align}  
where $f(\phi_k,\phi_j,\mathbf{X})$ represents the data-dependent pairwise relations between two latent nodes, and $\phi_k, \phi_j$ are the parameters for node $k$ and $j$ respectively. 

Step 3. \textit{Latent-to-visible propagation.} We update the conv-features by propagating the messages from the latent nodes back to the conv-feature nodes:
\begin{align}
\mathbf{\tilde{x}}_i &= h \left(\sum_{k=1}^d \psi(\mathbf{x}_i,\theta_k)\mathbf{\tilde{z}}_k\right), \quad 1\leq i \leq N,\label{eq:down}
\end{align} 
where $h$ is an element-wise activation function. And the output of the final step is used as an estimate of the non-local context for feature $\mathbf{x}_i$.

\subsection{An Equivalent Low-rank Kernel Representation}\label{subsec:kernel}

To gain more insight on the latent GNN model, we provide an alternative interpretation of our message passing algorithm in this subsection. We will show that our non-local GNN model is equivalent to using a low-rank representation for the affinity matrix in the fully-connected GNN model. To see this, we first represent our message passing algorithm in a matrix form. Specifically, denote $\bm{\psi}(\mathbf{x}_i) = [\psi(\mathbf{x}_i,\theta_1),\cdots, \psi(\mathbf{x}_i,\theta_d)]^\intercal$, we can write Eqn~\eqref{eq:up} in its matrix form as follows,
\begin{align}
\mathbf{Z} &= \bm{\Psi}(\mathbf{X})^\intercal\mathbf{X}\mathbf{W}\\
\bm{\Psi}(\mathbf{X}) &= [\bm{\psi}(\mathbf{x}_1),\cdots,\bm{\psi}(\mathbf{x}_N)]^\intercal \in\mathbb{R}^{N\times d}. 
\end{align}   
Similarly, let $\mathbf{F}_\mathbf{X} = [f(\phi_i,\phi_j,\mathbf{X})]_{d\times d}$, and the remaining two steps of the message passing can be written as:
\begin{align}
\mathbf{\tilde{Z}} &= \mathbf{F}_\mathbf{X} \mathbf{Z},\\
\mathbf{\tilde{X}} & = h\left(\bm{\Psi}(\mathbf{X})\mathbf{\tilde{Z}} \right).
\end{align}
Putting them together, we have the following matrix form for our message passing updating equations:
\begin{align}
\mathbf{\tilde{X}} & = h\left(\bm{\Psi}(\mathbf{X})\mathbf{F}_\mathbf{X} \bm{\Psi}(\mathbf{X})^\intercal\mathbf{X}\mathbf{W} \right)\label{eq:lgnn_mp}
\end{align}
Comparing the above equation with Eqn~\eqref{eq:matrix_mp}, we can see that our latent GNN can be viewed as a fully-connected graph neural network where
\begin{align}
\mathbf{A}(\mathbf{X})=\bm{\Psi}(\mathbf{X})\mathbf{F}_\mathbf{X} \bm{\Psi}(\mathbf{X})^\intercal,
\end{align}
which has a low-rank of $d$ instead of its full size $N$. 

\begin{figure}
	\centering
	\includegraphics[width=1.0\linewidth]{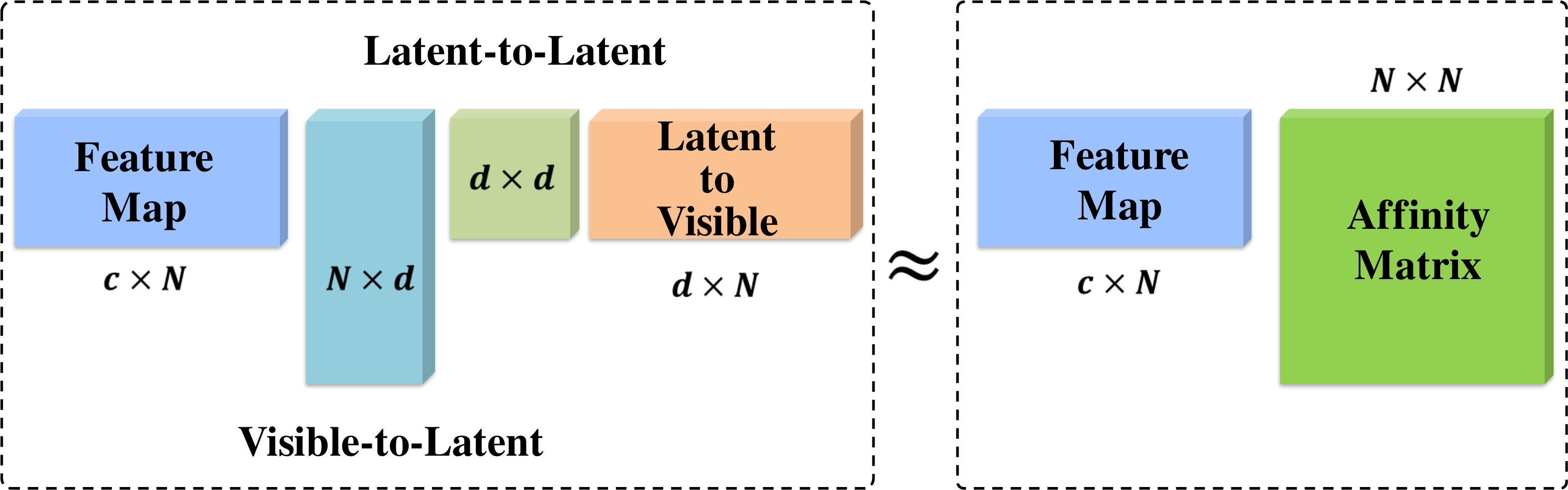}
	\caption{Computation diagram of the feature augmentation.}
	\label{fig:learn_normalize_adj}
\end{figure}

\paragraph{Fast Message Passing} One advantage of this low-rank representation is that it allows us to compute the matrix multiplications in Eqn~\eqref{eq:lgnn_mp} from right to left as in the message passing algorithm. This process does not require explicitly maintaining the full affinity matrix and significantly reduces the computation cost. Our method only has a linear complexity in terms of graph size, i.e., $\mathcal{O}(N\cdot c\cdot d)$ and achieves $N/d$ times speed-up compared to the non-local graph neural networks. 

\paragraph{Learning Low-rank Adjacency Matrix}

While the low-rank affinity matrix leads to significant improvements in computation efficiency, it has a limited representation power to model non-local relations between features. Inspired by the multiple kernel learning, we propose to extend the basic latent graph neural network by introducing a learnable mixture of low-rank matrices for capturing more complex non-local relations. 

Formally, we define a set of local affinity functions $\{\bm{\Psi}^{(m)}(\mathbf{X}),\mathbf{F}_\mathbf{X}^{(m)}\}$, which are used to construct the full affinity matrix as follows,
\begin{align}
\mathbf{A}(\mathbf{X})=\sum_m w_m \bm{\Psi}^{(m)}(\mathbf{X})\mathbf{F}_\mathbf{X}^{(m)} \bm{\Psi}^{(m)}(\mathbf{X})^{\intercal},
\end{align}  
where $w_m$ are the weights for each low-rank matrix. It is straightforward to extend the message passing algorithm to the case of mixture of low-rank kernels, and we can also jointly learn the mixture weights with the graph network.

\subsection{Feature Augmentation for Visual Recognition}\label{subsec:incorporate}

Given the context feature $\mathbf{\tilde{X}}$, we adopt a simple feature augmentation strategy to integrate the context representation with the original conv-features. Specifically, we use a weighted summation of two features as in the Residual Network~\cite{he2016deep}:
\begin{align}
		\mathbf{X}_{\text{aug}} &= \lambda \mathbf{\tilde{X}} + \mathbf{X}
\end{align}
where $\lambda$ is a scaling parameter.

As our latent graph neural network only takes a set of conv-features as input and is fully differentiable, it can be easily integrated with the existing deep ConvNets and applied to a variety of visual cognition tasks. In practice, we usually stack several graph convolutional layers into one feature augmentation module. In this work, we evaluate our feature augmentation module on several visual recognition tasks, including object detection and segmentation, and point cloud semantic segmentation. For object detection and instance segmentation, we insert our LatentGNN into the backbone network, at the end of each ResNet layer. For point cloud semantic segmentation, the LatentGNN module is inserted after each point feature extraction block.

\section{Experiments}\label{sec:experiments}
 
In this section, we conduct a series of experiments to validate the effectiveness of our method. We evaluate our model on three visual recognition tasks, including object detection and instance segmentation on the MSCOCO 2017 dataset~\cite{lin2014microsoft}, and point cloud semantic segmentation on the ScanNet dataset~\cite{dai2017scannet}. In each task, we compare our approach to the state-of-the-art non-local methods in terms of prediction accuracy and computational complexity. 

Below we present our experimental analysis on the MSCOCO dataset in Sec.~\ref{subsec:coco}, followed by our results on the ScanNet dataset in the Sec.~\ref{subsec:scanent}. In both tasks, we first introduce implementation details of our method, and then report the ablative study and comparisons of quantitative results on each dataset.

\begin{table*}[ht]
\center
\resizebox{0.85\textwidth}{!}{
\begin{tabular}{c| c|c|ccc|ccc|c|c}
\toprule
Model    &Stage&Kernels&$\text{AP}_{box}$ & $\text{AP}_{box}^{50}$ & $\text{AP}_{box}^{75}$ &  $\text{AP}_{sem}$ & $\text{AP}_{sem}^{50}$ & $\text{AP}_{sem}^{75}$ & FLOPS & \#Params\\ 			\midrule
ResNet-50\footnotemark[1] &-&-& 38.0 & 59.6 & 41.5 &  34.6 & 56.4 & 36.5 & - & -\\ 
 \shortstack[l]{\quad+NL Block\footnotemark[1]}
  			  & Stage4 & 1 & 39.0 &  61.1 & 41.9 & 35.5 &  58.0  &  37.4  & +10.67G & + 2.09M\\ \midrule\midrule
ResNet-50(1x)\footnotemark[2] & - &   -    & 37.8 &  59.1 & 41.2 & 34.2 &  55.8  &  36.3  & - &- \\ 
 \shortstack[l]{\quad+ NL Block\footnotemark[2]} 
			   & Stage4 & 1& 38.7 &  60.2 & 42.2 & 35.0 &  57.0  &  37.1  & +10.67G & + 2.09M \\ \hline
 \shortstack[l]{\quad+ $\text{LatentGNN}$}
			 & Stage3  & 1 & 38.2 &  59.7 & 41.7 & 34.7 &  56.3  &  36.8  & +1.48G & \textbf{+ 0.06M} \\ 
 \shortstack[l]{\quad+ $\text{LatentGNN}$}
			 & Stage4 & 1 & 39.0 &  60.7 & 42.6 & 35.2 &  57.6  &  37.4  & +1.11G & + 0.20M \\ 
 \shortstack[l]{\quad+ $\text{LatentGNN}$}
			 & Stage5 & 1 & 38.8 &  61.0 & 42.0 & 35.0 &  57.6  &  37.0  &  \textbf{+0.97G} & + 0.81M \\ 
 \shortstack[l]{\quad+ $\text{LatentGNN}$}
			 &Stage345& 1 &  \textbf{39.5} & \textbf{61.6} & \textbf{43.2}  & \textbf{35.6} & \textbf{58.3} & \textbf{37.7}&  +3.59G  &  +1.07M  \\  
\midrule
\midrule
ResNet-101(1x)& - &   -    & 39.9 &  61.3 & 43.8 & 35.9 &  58.2  &  38.1  & - & -\\  
\shortstack[l]{\quad+ $\text{LatentGNN}$}
			  & Stage4 & 1 & 41.0 &  63.2 & 45.0 & 36.9 &  59.6  &  39.4  &  \textbf{+1.11G} & \textbf{+ 0.20M}\\ 
\shortstack[l]{\quad+ $\text{LatentGNN}$}
			 &Stage345 & 1 & \textbf{41.4} & \textbf{63.7} & \textbf{45.2}& \textbf{37.2} & \textbf{60.1} & \textbf{39.5}  & +3.59G  &  +1.07M   \\  
\midrule
\midrule
ResNeXt-101(1x)& - &  -    & 42.1 & 64.1 &  45.9 & 37.8 &  60.3  & 39.5 & - & -\\
\shortstack[l]{\quad+ $\text{LatentGNN}$}
			  & Stage4 & 1 & 43.0 &  65.3 & 46.9 & 38.5 &  61.9  & 40.9 &  \textbf{+1.11G} & \textbf{+ 0.20M}\\ 
 \shortstack[l]{\quad+ $\text{LatentGNN}$}
			  &Stage345& 1 & \textbf{43.2} & \textbf{65.6} &\textbf{47.2} & \textbf{38.8} & \textbf{62.1} & \textbf{41.0}  &  +3.59G & +1.07M  \\  

 \bottomrule

\end{tabular}}
\caption{Performance of Non-local NNs and LatentGNNs with different \textbf{backbone} and different \textbf{stages} on MSCOCO 2017 validation set. All experiments are conducted with 1x learning rate schedule as in~\cite{he2017mask}.}
\label{tab:table_variants}
\end{table*}

\subsection{Object Detection \& Segmentation on MSCOCO}\label{subsec:coco}

\subsubsection{Network Architecture}
To utilize our LatentGNN in the task of object detection and instance segmentation, we adopt the standard setup of MaskRCNN~\cite{he2017mask}, in which ResNet/ResNeXt with FPN are used as the backbone architecture. We integrate our network module with the backbone network at different stages in the following experiments, and denote them as \textbf{+LatentGNN}. For ResNet backbones, we insert a LatentGNN into different residual blocks (Res3, Res4 or Res5) and refer them to as stage-3,4,5 respectively.     

To build a basic LatentGNN module with a single kernel, we first reduce the channel dimension of feature map with a $1\times 1$ convolutional bottleneck layer. We then use a learnable graph adjacency matrix for the message propagation from visible to latent nodes and in the reverse direction, which are implemented by a linear layer followed by a non-linear activation function (e.g. ReLU). The number of latent nodes $d$ is different for each stage of the backbone network due to different resolution of the feature maps and we set $d=\{150,100,50\}$ for the stage-3, stage-4 and stage-5, respectively. 
Moreover, we extend the basic LatentGNN module by introducing a learnable low-rank adjacency matrix with multiple kernels. In the multi-kernel LatentGNN, we aggregate the multiple visible features generated by the \textit{latent-to-visible propagation}, followed by a skip-connection to fuse them with the original conv feature.

For fair comparison, in addition to reporting the results in~\cite{wang2018non}, we implement a Non-local Neural Network using the same backbone network and graph affinity matrix $A_{sim}$. Following~\cite{wang2018non}, we insert the non-local block on the top of the last stage in the backbone network, which is denoted as '+NL Block'. 


\subsubsection{Training Setting}
We train our models for object detection and instance segmentation on the MSCOCO dataset and utilize the open-source implementation~\cite{massa2018mrcnn}. MSCOCO dataset is a challenging dataset that contains 115K images over 80 categories for training, 5K for validation and 20K for testing. Following the training procedure in \cite{massa2018mrcnn}, we resize the images so that the shorter side is 800 pixels. All experiments are conducted on 8 GPUs with 2 images per GPU (effective minibatch size 16) for 90K iterations, with a learning rate of 0.02 which is decreased by 10 at the 60K and 80K iteration. We use a weight decay of 0.0001 and momentum of 0.9. For ResNeXt, we train the model with 1 image per GPU and the same number of iterations, with a starting learning rate of 0.01. The backbone of all models are trained on ImageNet classification~\cite{deng2009imagenet}.

\subsubsection{Quantitive Results and Analysis}
The ablation study is conducted on the MSCOCO validation set. All standard MSCOCO metrics including AP, AP$^{50}$, AP$^{75}$ for both bounding boxes and segmentation masks are reported below.
\paragraph{Different Stage}
We first investigate at which stage we should add the LatentGNN layer to augment the feature with contextual information. We insert one LatentGNN layer for different residual stages, based on the  ResNet-50 backbone, to build the network. The quantitative results shown in Table.\ref{tab:table_variants} demonstrate the effects of our LatentGNN layer for each of and all three stages. 
\footnotetext[1]{Results reported by \cite{wang2018non}.}
\footnotetext[2]{Our implementation based on~\cite{massa2018mrcnn}.}

We note that incorporating LatentGNN layer into the stage3 has a small gain compared with the baseline method as the lower-level feature maps lack semantic information. Meanwhile, the stage4 equipped with a single LatentGNN layer could achieve $1.2$ points improvement for box AP and $1.0$ improvements for segmentation AP. Moreover, it is straightforward to incorporate our LatentGNN layer into multiple residual stages in order to augment the feature maps with multiscale contextual information. While easy to implement in our framework due to its efficient computation, it is not applicable to the Non-local Neural Networks. Our results show that adding LatentGNN module to the last three stages(3,4 and 5) together can achieve $39.5$ for box mAP and $35.6$ for segmentation mAP, which has $1.7$ and $1.4$ performance gain, respectively. The quantitative results indicate that our LatentGNN can achieve a large margin improvement with less FLOPS and fewer learnable parameters compared to the standard non-local block.

\paragraph{Different Backbones}
We validate our method on different backbone networks, ranging from ResNet-50, ResNet-101 to ResNeXt-101~\cite{xie2017aggregated}. It is worth noting that even when adopting stronger
backbones, the gain of LatentGNN compared to the baseline is still significant, which demonstrates that our LatentGNN is complementary to the capacity of current models. Equipped with different backbone, we also follow the same strategy to insert the LatentGNN layer into each residual stage. With feature augmentation on the last three stages, we can achieve 1.5 $\text{AP}_{box}$ and 1.3 $\text{AP}_{box}$ improvement with ResNet-101, and 1.1 $\text{AP}_{box}$ and 1.0 $\text{AP}_{box}$ improvement with ResNeXt-101.

\paragraph{Mixture of Low-rank Adjacency Matrices}

Using single kernel usually has limited representation power as aforementioned, and hence we also investigate the LatentGNN with a mixture of low-rank matrices for capturing the complex non-local relations. We conduct an ablation study to explore the influence of multiple low-rank adjacency matrices and report the performance in Table.\ref{tab:table_branches}. The quantitative performance indicates the mixture model is more powerful to represent the semantic context than the single kernel, and using three low-rank adjacency matrices can achieve 1.4 $\text{AP}_{box}$ and 1.2 $\text{AP}_{box}$ gain compared to the baseline. However, we note that inserting such mixture LatentGNN module into multiple stages achieves little gain, possibly due to the global context encoded by the multi-stage model.

\begin{table*}[t]
\center
\resizebox{0.8\textwidth}{!}{
\begin{tabular}{c|c|c|ccc|ccc|c|c}
\toprule
Model  & Stage &Kernels & $\text{AP}_{box}$ & $\text{AP}_{box}^{50}$ & $\text{AP}_{box}^{75}$ &  $\text{AP}_{sem}$ & $\text{AP}_{sem}^{50}$ & $\text{AP}_{sem}^{75}$ & FLOPS & \#Params \\ 			\midrule
ResNet-50(1x)    & - & - &  37.8  &  59.1  & 41.2 & 34.2 &  55.8  & 36.3 & - &- \\ \midrule
\multirow{3}{*}{\shortstack[c]{\quad+LatentGNN }} 
  & Stage4 & 1 & 39.0 &  60.7 & 42.6 & 35.2 &  57.6  &  37.4  & \textbf{+1.11G} & + \textbf{0.20M}\\ 
  & Stage4 & 2 & 39.0 &  60.7 & 42.7 & 35.3 &  57.6  &  37.6  & +1.30G & + 0.29M\\
  & Stage4 & 3 & \textbf{39.2} &  \textbf{61.0} & \textbf{42.8} & \textbf{35.4} &  \textbf{57.6}  &  \textbf{37.7} & +1.48G & +0.38M  \\ 
 \midrule
 \multirow{2}{*}{\shortstack[c]{\quad+LatentGNN }} 

 &Stage345 & 1 & 39.5 & 61.6 & 43.2  & 35.6 & 58.3 & 37.7& \textbf{+3.59G} & \textbf{+1.07M} \\ 
 &Stage345 & 3 & \textbf{39.5} & \textbf{61.7} & \textbf{43.3}  & \textbf{35.7} & \textbf{58.4} & \textbf{37.8}&  +5.13G  & +1.89M \\  
\bottomrule
\end{tabular}}
\caption{Performance of LatentGNN with a mixture of low-rank adjacency matrices on the MSCOCO validation set.}
\label{tab:table_branches}
\end{table*}

\begin{table*}[t]
\center
\resizebox{0.85\textwidth}{!}{
\begin{tabular}{c|c|c|cc|cc|c|c}
\toprule
Model  & Kernels & Scale &\shortstack[c]{Pixel \\Accuracy}  & \shortstack[c]{Voxel \\Accuracy}  &\shortstack[c]{ Class Pixel \\  Accuracy}  &\shortstack[c]{ Class Voxel \\  Accuracy}   & FLOPS & \#Params\\    \midrule

3DCNN\cite{dai2017scannet} & - &- & - & 73.0 &-&-& -  & -\\
PointNet\cite{qi2017pointnet} & -& - & - & 73.9 &-&-& -  & -\\
PointCNN\cite{li2018pointcnn} & - & - & 85.1 &-&-& -  & -\\
\midrule

PointNet++\cite{qi2017pointnet++}& - & Single Scale& 81.5 & 83.2   & 51.7  & 53.1& -  & - \\ 
PointNet++\cite{qi2017pointnet++}& - &Multi Scale & - & 84.5   & -  & -& -  & - \\ \midrule
\shortstack[c]{\quad+NL Block}
  & 1 & Single Scale & 82.3 & 84.0  &  53.1 &  54.5  & +31M & +0.70M  \\ 
\shortstack[c]{\quad+LatentGNN} 
  & 1 &Single Scale & 82.6 &  84.2 & 53.2  &  54.6 & \textbf{+15M} & \textbf{+0.31M}\\ 
\shortstack[c]{\quad+LatentGNN} 
  & 3 &Single Scale& \textbf{83.7} & \textbf{85.2}  &  \textbf{56.0} & \textbf{57.6}& +30M & +0.54M \\
 
\bottomrule
\end{tabular}}
\caption{Performance of LatentGNN on \textbf{ScanNet} based on PointNet++.}
\label{tab:table_scannet}
\end{table*}

\subsection{Point Cloud Segmentation on ScanNet}\label{subsec:scanent}
\subsubsection{Network Architecture}
We also validate our method on non-grid data, and conduct 3D point cloud semantic segmentation experiments on the ScanNet dataset, a 3D indoor scene benchmark with 1513 scanned and reconstructed indoor scenes in total. We follow the PointNet++ framework~\cite{qi2017pointnet++}, which introduces a hierarchical neural network to learn multi-scale feature descriptor for the point cloud. The PointNet++ uses a \textit{set abstraction module} (SA) as encoder and a \textit{feature propogation module} (FP) as decoder, which implement downsampling and upsampling of point cloud respectively. In particular, the {set abstraction module} (SA) learns local feature with increasing contextual scales and generates semantic feature of the point cloud. Hence, we insert the LatentGNN module at the end of each set abstraction layer to augment the feature with long-range dependencies, with latent dimension of \{80,40,20,10\} for four set abstraction layers, respectively.

\subsubsection{Training Setting}
We implement a multi-GPU version of the PointNet++ and conduct a series of experiments on ScanNet dataset for the task of semantic segmentation, which is to predict semantic object label for points in indoor scans. We use 1201 scenes for training and 312 scenes for test. We remove RGB information in all our experiments and convert point cloud label prediction into voxel labeling following~\cite{dai2017scannet}. For training, we take the preprocessed ScanNet data with 8192 points per data chunk. Specifically, all models are trained with a batch-size at 64 on 4 GPUs and the mini-batch size is set to 16 per GPU. We train our models for 200 epochs with Adam as optimizer, starting from the learning rate at 0.001 and reduce it using exponential schedule with the decay rate at 0.7.  We report the average voxel classification accuracy as the evaluation metric.

\subsubsection{Results}
We compare our model with the PointNet++ baseline and the Non-local Network module, and the performance is reported in Table.~\ref{tab:table_scannet}. 
Our proposed mixture LatentGNN model achieves the best performance in terms of per-voxel accuracy. The quantitative results show our LatentGNN achieves a large-margin performance gain at $2\%$ on voxel accuracy compared with PointNet++. We note that we do not use the multi-scale grouping(MSG) in \textit{set abstraction module}, which addresses the issue of non-uniform sampling density and may further improve the performance. Our single-kernel LatentGNN module also outperforms the Non-local Network module with much fewer parameters and merely half of computation cost. The results demonstrate the advantage of our efficient feature augmentation module.

\section{Conclusion}~\label{sec:con}
In this paper, we have presented an efficient and flexible context-aware feature augmentation model for visual recognition tasks. We introduce a novel graph neural network with a set of latent nodes to capture long-range dependencies. The sparse structure of this graph enable us to develop a scalable and efficient message passing algorithm for computing the context features. In addition, by introducing a grouping structure in the latent nodes, we are able to build a rich model of pairwise relations. We have evaluated our method on grid-like conv-feature maps for object localization and irregular point clouds for semantic segmentation. Our results demonstrate that our approach outperforms the baseline and other non-local method, and in particular, with significant lower computation cost in model inference. 



\clearpage
\section*{Acknowledgement}
The authors would like to acknowledge the support of a NSFC Grant No. 61703195 and a NSF Shanghai Grant No. 18ZR1425100.
\bibliography{egbib}
\bibliographystyle{icml2019}


%
%

\end{document}